# Rules, Belief Functions and Default Logic*

Nic Wilson
*Department of Computer Science*
*Queen Mary and Westfield College*
*Mile End Rd., London E1 4NS, UK*

## Abstract

*This paper describes a natural framework for rules, based on belief functions, which includes a representation of numerical rules, default rules and rules allowing and rules not allowing contraposition. In particular it justifies the use of the Dempster-Shafer Theory for representing a particular class of rules, Belief calculated being a lower probability given certain independence assumptions on an underlying space. It shows how a belief function framework can be generalised to other logics, including a general Monte-Carlo algorithm for calculating belief, and how a version of Reiter's Default Logic can be seen as a limiting case of a belief function formalism.*

## 1. Introduction

Rules used by people are often not completely reliable so any attempt to represent them must cope with the conclusion of the rule sometimes being incorrect. Numerical approaches do this by giving some kind of weighting to the conclusion of an uncertain rule; non-monotonic reasoning, a symbolic approach, ensures that these rules are defeasible, so that their conclusions could later be retracted if necessary.

There has been little work done, however, on relating numerical and symbolic techniques, an exception being the work of Adams [Adams, 66] further developed by Geffner and Pearl [Geffner, 89; Pearl, 88] where a logic is produced from probability theory, by tending the probabilities to 1.

This paper shows how a belief function approach can represent numerical rules, both those allowing contraposition and those not allowing contraposition, and how default rules may be viewed as the limiting case of such rules, when the certainty of the rule tends to 1. This allows the integration of the Dempster-Shafer Theory (DST) [Shafer, 76] and Reiter's Default Logic [Reiter, 80], hence enhancing the understanding of both.

Section 2 deals with the representation of numerical rules within DST, 2.1 giving an interpretation of the type of rule that DST typically represents; 2.2 presents a belief function framework that allows the theory to be generalised to other logics, and 2.3 shows how the framework can be applied to include rules which don't allow contraposition. Section 3 deals with the representation of default rules within the framework: 3.1 reformulates Reiter's Default Logic and defines a modified extension (equivalent to Lukaszewicz's); 3.2 shows how the belief function framework can be turned into a logic and 3.3 shows how to represent default rules within this logic. Section 4 indicates how priorities between rules can be represented, and Section 5 suggests how numerical and default rules could be used together within the framework.

## 2. Numerical Rules

Expert Systems like MYCIN [Buchanan and Shortliffe, 84] use uncertain rules of the form **If $a$ then $c$** : $(\alpha)$, where $\alpha$ is the some measure of how reliable the rule is. There are many ways of interpreting such a rule. We consider a natural interpretation which leads to the standard Dempster-Shafer representation of rules.

### 2.1 Justifying DST Representation of Rules

The standard way of representing the rules **If $a_i$ then $c_i$** : $(\alpha_i)$ (for $i = 1, \ldots, m$) with the Dempster-Shafer Theory is, for each rule to produce a simple support function with mass $\alpha_i$ allocated to the material implication $a_i \to c_i$ and the remaining mass $1 - \alpha_i$ allocated to the tautology, and then to combine these simple support functions by repeated application of Dempster's Rule. Pearl has criticised this representation for its behaviour under chaining and reasoning by cases. However it turns out that this DST approach represents a very natural type of rule.

The uncertain rule may in fact be an approximation to the certain rule $n \wedge a \to c$ where $n$ is an unknown antecedent or one too complicated to be easily expressed by the expert but which they judge to be true with probability $\alpha$. After all '... uncertainty measures characterise invisible facts, i.e., exceptions not covered in the formulas' [Pearl, 88, p2].

Since $n \wedge a \to c$ is logically equivalent to $n \wedge \neg c \to \neg a$, such a rule allows contraposition, and since it's also logically equivalent to $n \to (a \to c)$, this rule may also be interpreted

*In a proportion $\alpha$ of worlds (or situations) we know the material implication $a \to c$ is true.*

If we represent such a rule by a simple support function, as described above, Belief is just the probability that we know $a \rightarrow c$ to be true, so that it's a lower probability for $a \rightarrow c$. Similarly if we have a number of such rules, $a_i \rightarrow c_i$ ($i = 1, \ldots, m$), represent them as simple support functions and combine these with Dempster's Rule, Belief is a lower probability, given certain independence assumptions on the $n_i$s.

Consider now the typical Reasoning by Cases situation: we're given two rules **If** $a$ **then** $c$ : ($\alpha_1$) and **If** $\neg a$ **then** $c$ : ($\alpha_2$) which we'll interpret as uncertain material implications $n_1 \wedge a \rightarrow c$ and $n_2 \wedge \neg a \rightarrow c$ with $\Pr(n_i) = \alpha_i$, $i = 1, 2$.

Pearl argues that any reasonable measure of belief should obey the Sandwich Principle: deducing from those two rules that belief in $c$ should be between $\alpha_1$ and $\alpha_2$; the Dempster-Shafer approach however gives that $\text{Bel}(c) = \alpha_1 \alpha_2$.

But it is clear why the Sandwich Principle is violated for this approach: knowing either $a$ or $\neg a$ increases our knowledge and hence our belief. In worlds where $n_1 \wedge \neg n_2$ is true, $c$ may be always false if $a$ is always false; in the event $\neg n_1 \wedge n_2$, $c$ may be always false if $a$ is always true, and in the event $\neg n_1 \wedge \neg n_2$ there is no constraint on $c$ so $c$ may again always be false. Only in the event $n_1 \wedge n_2$ can we be sure that $c$ is true, so making the assumption of independence of $n_1$ and $n_2$ (which is reasonable without contrary knowledge) we get $\text{Bel}(c)$, the probability that we're in a world where we know $c$ to be true, is $\alpha_1 \alpha_2$.

This type of rule can also be chained:
$n_1 \wedge a \rightarrow b$ and $n_2 \wedge b \rightarrow c$ with $\Pr(n_i) = \alpha_i$, ($i = 1, 2$) leads to $(n_1 \wedge n_2) \wedge a \rightarrow c$, and again assuming independence of $n_1$ and $n_2$ this gives $\Pr(n_1 \wedge n_2) = \alpha_1 \alpha_2$. If we now learn that $a$ is true we get $\Pr(c) \geq \alpha_1 \alpha_2$ and so $\text{Bel}(c) = \alpha_1 \alpha_2$, the result given by application of Dempster's Rule.

Of course the assumption of independence of the $n_i$s will not always be valid—if correlations between the rules are known they should be (and can be) incorporated.

The Dempster-Shafer approach is thus a natural, formally justified as well as a computationally efficient way (see [Wilson, 89] and section 2.2) to represent If-Then rules.

### 2.2 The Sources of Evidence Framework

We will be interested in extending DST to other logics (see [Saffioti, 90] for other work on this, and see [Ruspini, 87] for a justification of DST using a modal logic of knowledge). A natural way to extend Dempster's multi-valued mapping [Dempster, 67] is as follows:

We have a mutually exclusive and exhaustive set $\Omega$ with a probability function P on it, and we're interested in the truth of formulae in $L$, where $L$ is the language of some logic. With each $\eta \in \Omega$ is associated a set $K_\eta$ ($\subseteq L$): the set of all formulae *known* to be true given that $\eta$ is true. For a formula $d \in L$, $\text{Bel}(d)$ is defined to be the probability that we know $d$ to be true i.e.,

$$\text{Bel}(d) = \sum_{\eta \,:\, \eta \Rightarrow d} P(\eta) = \sum_{\eta \,:\, K_\eta \ni d} P(\eta).$$

Justifying Dempster's Rule for general belief functions is problematic* so we restrict ourselves to the combination of a finite number of simple support functions and, to justify this, use the Sources of Evidence framework (based on Shafer's random sources canonical example [Shafer, 87]; see [Wilson, 89] for details).

Suppose we have distinct propositions $n_i$, $i = 1, \ldots, m$, (not in $L$) and for each we have a prior probability $\alpha_i$. Suppose also that we know that if $n_i$ is true, some evidence $\text{Evd}_i$ is also true, where $\text{Evd}_i$ is a statement about the logic (it might for example be that the material implication $a_i \rightarrow c_i$ is true, as in section 2.1). If $n_i$ is not true we know nothing about the truth of $\text{Evd}_i$.

We also allow there to be a set of facts $W$ which are known certainly to be true.

$n_i$ may, as the name of the framework suggests, represent the event that a source of evidence, which tells us evidence $\text{Evd}_i$, is reliable. Alternatively $n_i$ may be an unknown antecedent of a rule, as described in the last section; or $n_i$ may just be some event for which, when it occurs, we are sure that the evidence $\text{Evd}_i$ is true.

Let $\eta_\sigma$ be the elementary event

$$\bigwedge_{i \in \sigma} n_i \wedge \bigwedge_{i \notin \sigma} \neg n_i$$

and let $\Omega$ be the mutually exclusive and exhaustive set of elementary events $\{\eta_\sigma : \sigma \subseteq \{1, \ldots m\}\}$. Take some probability function P on $\Omega$.

---

\* For example there is the problem of the collapsing of the Belief-Plausibility interval, e.g., [Pearl, 89]; see also Shafer's presentations of his random codes canonical example, with discussion [Shafer, 82a, 82b].

If we think of P as saying, for each $\eta_\sigma$, the probability that we are in a world in which $\eta_\sigma$ is true, then Bel($d$) is the probability that we are in a world in which we know $d$ is true.

Bel maybe viewed as a lower probability given the probability function P on the underlying space $\Omega$.

It is argued in [Wilson, 89] that, in the absence of correlation information on the $n_i$s, certain assumptions (A) and (B) are entirely reasonable. (A) is roughly that, since an unreliable source/rule doesn't give us any information, it shouldn't affect the probabilities (an example of the application of this assumption is given below in 2.3); (B) is that, if the sources are not contradictory (i.e., $\bigwedge_1^m n_i$ is not known to be impossible) then we take, for each $i$, $P(n_i)$ to be $\alpha_i$, its prior value. These assumptions determine a unique probability function $P^{DS}$ given by

$$P^{DS}(\varepsilon_\sigma) = \begin{cases} 0, & \text{if } K_\sigma \text{ inconsistent,} \\ \rho_\sigma/k, & \text{otherwise,} \end{cases}$$

$$\text{where} \quad k = \sum_{K_\sigma \text{consistent}} \rho_\sigma$$

$$\text{and} \quad \rho_\sigma = \prod_{i \in \sigma} \alpha_i \prod_{i \notin \sigma}(1 - \alpha_i).$$

This is, in fact, the probability function that leads to Dempster-Shafer belief when each $Evd_i$ is that some proposition $p_i$ is true: the belief as defined above will be the same as that calculated by using Dempster's Rule to combine simple support functions with mass $\alpha_i$ attributed to the proposition $p_i$, $i = 1, \ldots, m$.

Since Belief, as defined here, is just 'randomised logic' the calculation of Belief inherits its computational efficiency from that of the underlying logic: Bel($d$) can be calculated, using the following Monte-Carlo algorithm:

*For each trial:-*

(i) **Pick** $\sigma$ with probability $P(\eta_\sigma)$

(ii) **If** $K_\sigma \ni d$ **then** trial succeeds else trial fails.

The proportion of successful trials then converges to Bel($d$).

Given that $P(\eta_\sigma)$ is not too hard to calculate, the calculation takes time proportional to the time it takes to check if $d \in K_\sigma$, but with a fairly large constant term corresponding to the number of trials needed to get reasonable accuracy.

With the probability function $P = P^{DS}$, step (i) can be performed very easily. Since any sensible measure of belief should collapse to the logic for the extreme case, its computational efficiency cannot hope to be better than that of the underlying logic. Thus the calculation of Dempster-Shafer belief is as fast, up to a constant, as the calculation of a measure of belief could possibly be. In particular it is shown in [Wilson, 89] that (up to arbitrary accuracy) Dempster-Shafer Belief on a mutually exclusive and exhaustive frame of discernment can be calculated in time approximately linear in the number of evidences and size of the frame of discernment.

### 2.3 Rules Not Allowing Contraposition

Some rules do not allow contraposition. For example the rule *Typically males don't have long beards* seems reasonable, and even mildly informative, but on meeting someone with a long beard, it would be unreasonable to deduce that they were female. In order to represent rules not allowing contraposition, inference rules such as $a\,/\,c$ will be used which, like the rules used in many Expert Systems, given $a$, allow the deduction of $c$, but given $\neg c$, do not allow $\neg a$ to be deduced.

Suppose we have a set of rules **If $a_i$ then $c_i$** : $(\alpha_i)$ for $i = 1, \ldots, m$, ($a_i$s and $c_i$s closed wffs in first order logic) for which we do not wish to allow contraposition. Let $I = \{a_i\,/\,c_i\,:\,i = 1, \ldots, m\}$ where the (certain) inference rule $a_i\,/\,c_i$ means 'if we know $a_i$ we can deduce $c_i$', and let $I_\sigma = \{a_i\,/\,c_i\,:\,i \in \sigma\}$.

For some set $U$ of closed wffs and set of inference rules $J$ we define $\text{Th}^J(U)$ to be the logical closure of $U$ when all the inference rules in $J$ are added to the logic i.e., the set of formulae obtained by applying all the inference rules in $J$ repeatedly to $U$, so that $\text{Th}^J(U)$ is the smallest set $\Gamma$ such that

(i) $\Gamma \supseteq U$,

(ii) $\text{Th}(\Gamma) = \Gamma$ and

(iii) if $a\,/\,c \in J$ and $a \in \Gamma$ then $c \in \Gamma$,

where $\text{Th}(\Gamma)$ means the logical closure of $\Gamma$ within first order logic.

Abbreviate $\text{Th}^{I_\sigma}(W)$ to $\text{Th}^\sigma(W)$.

To represent this set of rules within the sources of evidence framework we make the $i$th evidence be that the inference rule $a_i\,/\,c_i$ is added to the logic (so that whenever $a_i$ is known, $c_i$ may be deduced). To be precise, we set $K_\sigma$ to be $\text{Th}^\sigma(W)$.

This includes the uncertain material implications, described in 2.1, as a special case: make, for all $i$, the $i$th inference rule equal $\top\,/\,(a_i \to c_i)$.

## Example

Whilst attempting to deduce information about our acquaintance Nixon we learn that he is a quaker and a republican, so that $W = \{\text{quaker}, \text{republican}\}$. Two rules, **If quaker then pacifist** : $(\alpha_1)$ and **If republican then ¬pacifist** : $(\alpha_2)$ are also known.

To represent these we take $\text{Evd}_1$ to be that the first rule is correct and that the corresponding inference rule **quaker / pacifist** should be added to the logic, and similarly for $\text{Evd}_2$.

Thus if $n_1$ then, if at any time we learn **quaker**, we will deduce **pacifist**. This gives

$$K_\emptyset = \text{Th}(W) = \text{Th}(\{\text{quaker}, \text{republican}\})$$
$$K_{\{1\}} = \text{Th}(W \cup \{\text{pacifist}\})$$
$$K_{\{2\}} = \text{Th}(W \cup \{\neg\text{pacifist}\})$$
$$K_{\{1,2\}} = \text{Th}(W \cup \{\text{pacifist}, \neg\text{pacifist}\}).$$

Since $K_{\{1,2\}}$ is inconsistent, $P(n_1 \wedge n_2)$ must be 0. In order to come up with a probability function P we make certain independence assumptions. Knowing only about one rule, the first, we would obviously take $P(n_1) = \alpha_1$; adding an unreliable second rule doesn't give us any information so shouldn't change this probability i.e., we make the assumption that $P(n_1|\neg n_2) = \alpha_1$. Symmetrically we make the assumption $P(n_2|\neg n_1) = \alpha_2$. Both these assumptions are instances of assumption (A) mentioned above in 2.2. Only in worlds when $n_1 \wedge \neg n_2$ is true (when $\sigma = \{1\}$) do we know **pacifist**, and only in worlds $\neg n_1 \wedge n_2$ ($\sigma = \{2\}$) do we know **¬pacifist**, so

$$\text{Bel}(\text{pacifist}) = P(n_1 \wedge \neg n_2) = \frac{\alpha_1(1 - \alpha_2)}{1 - \alpha_1\alpha_2}$$
$$\text{Bel}(\neg\text{pacifist}) = P(\neg n_1 \wedge n_2) = \frac{(1 - \alpha_1)\alpha_2}{1 - \alpha_1\alpha_2}.$$

If the reliabilities of the two rules are the same then $\text{Bel}(\text{pacifist}) = \text{Bel}(\neg\text{pacifist})$. If, on the other hand, the first rule is very reliable, but the second isn't so reliable then $\text{Bel}(\text{pacifist})$ will be close to 1 and $\text{Bel}(\neg\text{pacifist})$ will be close to 0.

## 3. Default Rules

An alternative to rules with numerical uncertainty are default rules—in the absence of information indicating that the circumstances are exceptional, the rule is fired, though the consequence of the rule may later have to be retracted, if it's discovered that circumstances are in fact exceptional.

### 3.1 Default Logic

Reiter's Default Logic [Reiter, 80] is a logic for reasoning with default rules. A default rule is a rule of the form 'If we know $a$ then deduce $c$, as long as $b$ is consistent', or $a : b / c$ for short.

Let $\Delta = (D, W)$ be a closed default theory where $L$ is the language of a first order logic, $W \subseteq L$, a set of closed wffs, are the facts and $D$ is the set of default rules
$$\left\{ \frac{a_i : b_i}{c_i} : i = 1, \ldots, m \right\}$$
where $a_i$, $b_i$ and $c_i$ are closed formulae, for each $i$.

It turns out that Reiter's default logic can be expressed in terms of inference rules. Let $I = \{a_i / c_i : i = 1, \ldots, m\}$. The behaviour of the defaults in $D$ will be mimicked by use of the corresponding inference rule in $I$.

Let $S = \{\text{Th}^\gamma(W) : \gamma \subseteq \{1, \ldots m\}\}$. $S$ contains all the sets of formulae produced by applying different subsets of the inference rules to $W$.

For some $K \in S$ an inference rule $a_i / c_i$ may have been applied even though $b_i$ is inconsistent (i.e., $\neg b_i \in K$), in which case the inference rule was not behaving like the corresponding Default rule. Then we say that $K$ is $\Delta$-inconsistent. Formally this property can be defined as follows:

$K \in S$ is $\Delta$-consistent if and only if there exists a $\gamma \subseteq \{1, \ldots m\}$ with $K = \text{Th}^\gamma(W)$ and $K \not\ni \neg b_i$ for all $i \in \gamma$.

In default logic the extensions are intended to be the different possible completions, using the default rules, of an incomplete set of facts about the world.

**Theorem 1:** $E$ is an extension if and only if $E = \text{Th}^\gamma(W)$ where $\gamma = \{i : \neg b_i \notin E\}$.

This shows that extensions are $\Delta$-consistent sets in $S$. In fact we have

**Theorem 2:** If $E$ is an extension of $\Delta$ then $E$ is a maximal $\Delta$-consistent set in $S$.

**Definition:** $E$ is said to be an M-extension of $\Delta$ if $E$ is a maximal $\Delta$-consistent set in $S$.

M-extensions are formed by applying as many inference rules as possible without contradicting $\Delta$-consistency. Theorem 2 showed that extensions are always M-extensions.

**Theorem 3:** Let $\Delta$ be a closed normal default theory. Then $E$ is an extension of $\Delta$ if and only if $E$ is a maximally $\Delta$-consistent set in $\mathcal{S}$.

Thus for closed normal default theories $E$ is an extension if and only if $E$ is an M-extension.

M-extensions have for general closed default theories the nice properties extensions only have for closed normal default theories:

**Theorem 4:** Every closed default theory has an M-extension.

**Theorem 5 (Semi-monotonicity):** Let $\Delta = (D, W)$, $\Delta' = (D', W)$ be closed default theories with $D \subseteq D'$. If $E$ is an M-extension of $\Delta$ then there exists an M-extension $E'$ of $\Delta'$ with $E' \supseteq E$.

We can also define an M-default proof, in an obvious way, which is complete, that is for any closed wff $p$ there is an M-default proof of $p$ if and only if $p \in E$ for some M-extension $E$.

It might be suggested that any M-extension of a default theory which is not an extension is not a sensible completion of one's knowledge: this however is not the case e.g., there are apparently coherent default theories that allow no extension (see [Wilson, 90] for an example, and also for proofs of the above results) but which, by Theorem 4, allow M-extensions.

M-extensions turn out to be the modified extensions defined in [Lukaszewicz, 84] (also see [Besnard, 89]).

### 3.2 The Sources of Evidence Framework as a Logic

To turn the Sources of Evidence Framework into a Logic, we tend the reliabilities of the sources (the $\alpha_i$s) to 1. B-extensions are the sets of formulae whose belief can be made to tend to 1. We can consider Bel as a function $\text{Bel}(\underline{\alpha}, p)$ where $\underline{\alpha} = (\alpha_1, \alpha_2, \ldots, \alpha_m)$ is the vector consisting of the reliabilities of all the sources.

To use the Sources of Evidence framework to produce a logic we require that for any closed wff $p$, $\text{Bel}(p)$ tends to either 0 or 1. A B-extension is then the set of formulae whose belief tends to 1.

Formally $E$ is a B-extension if and only if for $i = 1, \ldots, m$ there exist monotonic functions $\alpha_i : [1, \infty) \longrightarrow [0, 1)$ with $\alpha_i(x)$ tending to 1 as $x$ tends to infinity, and given $\epsilon > 0$ there exists $N_\epsilon$ such that for all $x > N_\epsilon$ and for all $p \in L$

$$\text{Bel}(\underline{\alpha}(x), p) > 1 - \epsilon \quad \text{if } p \in E$$
$$\text{Bel}(\underline{\alpha}(x), p) < \epsilon \quad \text{if } p \notin E.$$

### Example Continued

In the case of Nixon we have 2 B-extensions. When the reliability of the first rule tends to 1 much faster than that of the second rule we get that $\text{Bel}(\text{pacifist})$ tends to 1, and $\text{Bel}(\neg\text{pacifist})$ tends to 0 so $K_{\{1\}} = \text{Th}(W \cup \{\text{pacifist}\})$ is a B-extension.
Similarly $K_{\{2\}} = \text{Th}(W \cup \{\neg\text{pacifist}\})$ is a B-extension.

**Theorem 6:** If $E$ is a B-extension then $E = K_\sigma$ for some $\sigma$.

If we don't have information about correlations between the sources we can reasonably make assumptions (A) and (B) giving $\text{P} = \text{P}^{\text{DS}}$.

**Theorem 7:** With $\text{P} = \text{P}^{\text{DS}}$, $E$ is a B-extension if and only if $E = K_\sigma$ for some $\sigma$ maximal with $K_\sigma$ consistent.

### 3.3 Representation of Default Rules in Sources of Evidence Framework

Default rules will be represented in the Sources of Evidence framework by treating them rather like numerical rules with a high, but unknown, certainty: roughly speaking we make the $i$th evidence $\text{Evd}_i$ be that the inference rule $a_i / c_i$ is a correct rule, as we did in 2.3, and take the limit as the reliabilities of the sources (that is, the certainties of the rules) tend to 1, to produce the B-extensions.

In the example we found that the B-extensions were just the same as Reiter's extensions. This was no coincidence: when the probability function $\text{P}^{\text{DS}}$ on $\Omega$ is used the B-extensions are exactly the M-extensions of the default theory.

#### 3.3.1 Closed Normal Default Theories

Let $\Delta$ be a closed normal default theory. We want the $i$th evidence to be that the inference rule $a_i / c_i$ is a correct rule, so, formally, we set $K_\sigma = \text{Th}^\sigma(W)$, and also set $\text{P} = \text{P}^{\text{DS}}$.

**Theorem 8:** Let $\Delta$ be a closed normal default theory. With the above representation of Closed Normal Default rules within the Sources of Evidence framework $E$ is a B-extension $\iff$ $E$ is an M-extension of $\Delta$ $\iff$ $E$ is an extension of $\Delta$.

### 3.3.2 General Closed Default Theories

We again set $P = P^{DS}$. To represent the consistency condition $b_i$ we have to be a little trickier. We first add new distinct symbols $q_1, \ldots, q_m$ to the alphabet of the language to get a new language $L'$.

We want the statement of the $i$th source to be that inferences rules $a_i / q_i$, $q_i / c_i$, $\neg b_i / \neg q_i$ are correct rules. The idea is that knowing $a_i$ will enable us to deduce $c_i$ unless $\neg b_i$ is known, in which case we will get an inconsistency since we'll know both $q_i$ and $\neg q_i$.

To be precise we let $K'_\sigma \subseteq L'$ be the theorems of $W$ when the inference rules

$$J_\sigma = \{\frac{a_i}{q_i},\ \frac{q_i}{c_i},\ \frac{\neg b_i}{\neg q_i}\ :\ i \in \sigma\}$$

are added to the logic, that is, $K'_\sigma \stackrel{\text{def}}{=} \text{Th}^{J_\sigma}(W)$.

We're only interested in the wffs in $L$ so let $K_\sigma$ be $K'_\sigma \cap L$, so that $K_\sigma$ is $K'_\sigma$ stripped of all formulae mentioning some $q_i$.

This gives the following result:

**Theorem 9:** With the representation of Default rules in the Sources of Evidence framework described above, for any set of closed wffs $E$

$E$ is a B-extension if and only if $E$ is an M-extension.

### 4. Expressing Preferences between Rules

Suppose we have two rules, **If penguin then ¬flies :** ($\alpha_1$), and **If bird then flies :** ($\alpha_2$), and two facts, $W = \{\text{penguin}, \text{bird}\}$.

Expressing these as inference rules gives, as in the Nixon example,

$$\text{Bel}(\neg\text{flies}) = \frac{\alpha_1(1-\alpha_2)}{1 - \alpha_1\alpha_2},$$
$$\text{and}\quad \text{Bel}(\text{flies}) = \frac{(1-\alpha_1)\alpha_2}{1 - \alpha_1\alpha_2}$$

arising from assumptions $P(n_1|\neg n_2) = \alpha_1$, $P(n_2|\neg n_1) = \alpha_2$.

But since we know that penguins are a subclass of birds it seems that the first rule should override the second: if we only knew about the first rule we would get $\text{Bel}(\neg\text{flies}) = \alpha_1$ which should not be changed on learning the second rule.

The preference of some sets of rules over others are represented by making some different assumptions on the probability function P that reflect that preference.

In this case we specify $P(n_1) = \alpha_1$ (since $P(n_1)$ should be unaffected by the addition of a second rule), and $P(n_2|\neg n_1) = \alpha_2$ is still an intuitive assumption to make.

This gives $P(n_1 \wedge n_2) = 0$, since $K_{\{1,2\}}$ is inconsistent, $P(n_1 \wedge \neg n_2) = \alpha_1$, $P(\neg n_1 \wedge n_2) = (1-\alpha_1)\alpha_2$, $P(\neg n_1 \wedge \neg n_2) = (1-\alpha_1)(1-\alpha_2)$, and $\text{Bel}(\neg\text{flies}) = \alpha_1$, $\text{Bel}(\text{flies}) = (1-\alpha_1)\alpha_2$.

Here $\text{Th}(\{\text{penguin}, \text{bird}, \neg\text{flies}\})$ is the only B-extension.

### 5. Combining Numerical and Default Rules

It has been shown how the Sources of Evidence framework can represent either numerical rules, or, taking the limit as the reliabilities of the rules tend to 1, default rules. The next step is to combine both within this framework.

Suppose that our knowledge includes both default rules and numerical rules. First we represent both as evidences, which add an inference rule to the logic, in the sources of evidence framework (which includes the contrapositioning rules as a special case). We first consider only the default rules, and produce the B-extensions. Then we add the other rules/sources to get a belief function in each B-extension.

For $d \in L$, $\text{BEL}_*(d)$ is defined to be the minimum value of $\text{Bel}(d)$ over the extensions, a rather conservative measure; $\text{BEL}^*(d)$ is defined to be the maximum value of $\text{Bel}(d)$. If $\text{BEL}^*(d)$ is high this gives at least some reason for believing $d$: there is some combination of default rules which if correct lend high support to $d$. Some average of Bel over the extensions could also be a useful measure.

Another way of looking at this is to consider Bel as a function of the unknown, but high reliabilities $\underline{\alpha} = (\alpha_1, \ldots, \alpha_m)$. $\text{BEL}_*(d)$ is then $\inf_{\underline{\alpha} \to \underline{1}} \text{Bel}(\underline{\alpha}, p)$, and $\text{BEL}^*(d)$ is $\sup_{\underline{\alpha} \to \underline{1}} \text{Bel}(\underline{\alpha}, p)$.

### 6. Concluding Comments

We have given counter-arguments to some of Pearl's criticisms of the use of belief functions to represent rules and argued that the Dempster-Shafer Theory is a natural way to represent a type of If-Then rule.

If it is known that the rules are correlated, then the independence assumptions may well not be justified, and so a more general belief function approach such as the sources of evidence framework is needed to allow the dependencies between the rules to be incorporated in the underlying probability function P.

Dependencies must also be used, as described in section 4, to represent dominance of certain rules (or chains of rules) over others. We have shown that this belief function approach also enables the representation of default rules.

Another very natural type of rule **If a then c :** ($\alpha$) related to that described in 2.1 is where, again, there is an unknown antecedent $n$ with $n \wedge a \rightarrow c$, but instead of knowing the prior probability of $n$, we know the conditional probability $P(n|a) = \alpha$. With a number of such rules we can take, as before, Belief as a lower probability, tend the $\alpha_i$s to 1 and see which Beliefs tend to 1. This is effectively the approach taken by Adams, Geffner and Pearl. It would be interesting to explore whether progress can be made by making independence assumptions on the $n_i$s, as we did for the type of rule described in 2.1.

It is clear that there is no single correct way of representing numerical If-Then rules. Future research in this area should attempt to clarify what different types of numerical rule there are, and to represent them within a single framework.

## Acknowledgements

I am greatly indebted to Mike Clarke, for numerous useful and interesting discussions, and without whom this paper could not have been written. I have also enjoyed many productive conversations with Mike Hopkins and my colleagues on the DRUMS project.